# FenceMask: A Data Augmentation Approach for Pre-extracted Image Features

Pu Li    Xiangyang Li    Xiang Long

Peking University

Beijing University of Posts and Telecommunications

**Abstract.** We propose a novel data augmentation method named 'FenceMask' that exhibits outstanding performance in various computer vision tasks. It is based on the 'simulation of object occlusion' strategy, which aim to achieve the balance between object occlusion and information retention of the input data. By enhancing the sparsity and regularity of the occlusion block, our augmentation method overcome the difficulty of small object augmentation and notably improve performance over baselines. Sufficient experiments prove the performance of our method is better than other simulate object occlusion approaches. We tested it on CIFAR10, CIFAR100 and ImageNet datasets for Coarse-grained classification, COCO2017 and VisDrone datasets for detection, Oxford Flowers, Cornel Leaf and Stanford Dogs datasets for Fine-Grained Visual Categorization. Our method achieved significant performance improvement on Fine-Grained Visual Categorization task and VisDrone dataset.

## 1   Introduction

Recently, Deep Convolutional Neural Networks (CNNS) has made remarkable achievements in computer vision tasks, showing outstanding performance in image classification, object detection, semantic segmentation and other **typical computer vision tasks.**

Meanwhile, a number of derived tasks have become the focus of current research, including fine-grained visual categorization and small or dim object detection which is necessary for vision tasks in real scenes. With the development of computational power and network size, the parameters of convolutional neural networks have been significantly increased, which cause usual and severe over-fitting problem in tasks with insufficient data. Data augmentation can generate more useful data from existing data and restrain the over-fitting problem. Because of simple implementation and needlessness of additional data, data augmentation has become an indispensable part of deep neural network training.

Occlusion simulation means to simulate common object occlusion problems in reality by deleting part of information in images, including Random Erasing [1], CutOut [2], Hide and Seek [3] and GridMask [4]. Occlusion simulation is now widely used due to its prominent performance in industrial deployment and simple implementation.

But how to keep balance between object occlusion and information retention is still significance research point, especially in the complex reality scene, the object size has a great deal of uncertainty, randomly generated occlusion block can lead the most of the object become failure case, it not only have an positive effect on data augmentation, but also misguide model during training.

The previous occlusion block generating algorithm is mostly based on randomly generating include CutOut, Random Erasing and Hide and Seek. GridMask proposed that occlusion block generating algorithm should consider the problem of keeping the object intact while expanding the deleted area, we think it is key to generate occlusion block. We randomly selected from 100 images from Stanford Dogs and Visdrone date set which include lots of small object, and use Cutout, Random erasing, Hide and Seek and GridMask algorithm dispose these images respectively. After the statistics of failure case, we find that the current occlusion algorithm is awfully harmful to such datasets. Therefore, we designed a occlusion block algorithm that can maintain the data occlusion rate and have better retention effect on small objects in a specific datasets, meanwhile this method can also show normal performance improvement in a general dataset.

We believe that further sparsity and regularization of occlusion block are effective design ideas. More sparsity of occlusion block means to reduce the area of single occlusion block and increase the density of it. In the visual task that include many small objects, sparser occlusion block can save more information and prevent failure case. Regularization of occlusion block means that the generated occlusion block should conform to specific arrangement rules to prevent occlusion block aggregation, which caused lots of failure cases. In addition, we find that the continuity of the object is not important for train. Even if our data augmentation algorithm destroys the continuity of the target, experiment in Section 4 proved it has no special effect on the final training result.

Based on the above analysis, we designed a data augmentation method with both important sparsity and regularity as shown in **Fig.1**. Although this method breaks the continuity of objects in image, the damage could prove to be acceptable, as shown in Fig.1. FenceMask is a simple design, easy to implement, and will not generate any additional compute. It can be applied to all convolutional neural networks and various computer vision tasks, especially show particularly good performance in some datasets that including more small objects or features.

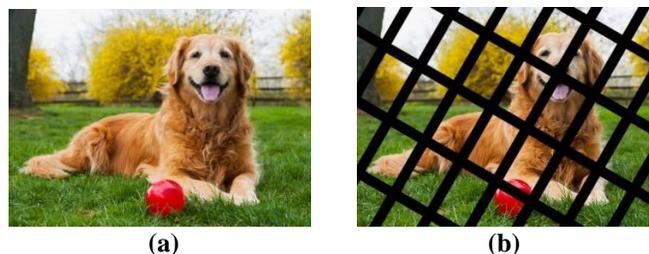

(a)          (b)
**Fig.1** Example of FenceMask data augmentation. (a) input image (b) FenceMask.

To prove the effectiveness of FenceMask, we designed the following experiment. In the image classification task, we tested our method on the CIFAR10, CIFAR100, ImageNet datasets. FenceMask demonstrated comparable capabilities with current occlusion approaches. Besides, we tested our method in the fine-grained detection task, the data disease detection task. We tested our method in the target detection task as well. First of all, we obtained the following performance in the COCO universal object detection dataset. Here's what the VisDrone achieves in its aerial photography dataset.

## 2 Related Work

### 2.1 Data Augmentation

Due to large-scale parameters, the neural network often occurs severe over-fitting, which weakens performance of the network. Data augmentation is a effective method to solve this universal problem. Compared with other solutions, data augmentation is easy implementation, only needs to handle input images instead of changing network structure or adding extra parameters. Thus, it almost does not increase additional computation, and can be applied to a variety of computer vision tasks. Data augmentation has become an indispensable step in neural network training.

Currently, there are four basic classes of data augmentation that are proved to improve performance of various visual tasks. Pixel-wise adjustment, information combination of multiple images, GAN based style transfer and simulation of occlusion simulation. Pixel-wise adjustment refers to the change of image information at the pixel level, mainly including photometric distortions. Photometric distortions refer to process image in color space, such as brightness, contrast, hue, saturation and add noise to image. Geometric distortions refer to process image in coordinate space, such as scaling, cropping, flipping and rotating. Information combination of multiple images can increase the information capacity of single input image by gathering information from multiple images including MixUp [5] and CutMix [6]. GAN [7] based style transfer can use CycleGAN to generate new images that have different style from original images. It can increase the richness of the scene and reduce the texture bias learned by CNN. Simulation of occlusion simulation can improve the perception ability to incomplete objects. For example, random erase and CutOut randomly select the rectangle region in an image and fill in a random or complementary value of zero. Hide-and-Seek divide the image into grids, and randomly select some rectangle regions. Grid mask randomly generate multiple rectangle regions, they regular arrange in an image. Similar concepts are also applied to feature maps, there are DropOut, DropConnect, and DropBlock approaches. Additionally, AutoAugment[8] improved the inception-preprocess using reinforcement learning to search existing policies for the optimal combination. Concretely, it creates a search space for data augmentation strategy and uses search algorithm to select the data augmentation strategy suitable for the specific dataset, this means that lots of computing power are consumed for searching.

### 2.2 Fine-Grained Visual Categorization

Fine-grained Visual Categorization is based on the distinction of basic categories, and carries out finer subclassification, such as distinguishing the species of birds, the style of cars, the breed of dogs, etc. At present, there are a wide range of business needs and application scenarios in industry and real life. Compared with coarse-grained images, fine-grained images have more similar appearance and characteristics, and in addition, the acquisition has the effects of attitude, perspective, light, occlusion, background interference, leading to the phenomenon of large inter-class differences and small intra-class differences in data presentation, thus making classification more difficult.

**Fig.2**. shows two species in the fine-grained image classification dataset CUB-200 [12], California gull and Arctic gull. From the comparison of the vertical images, it can be seen that the two different species are very similar in appearance, while from the contrast horizontal direction, it can be seen that the same species has large intra-class differences due to the different postures, backgrounds and shooting angles. Therefore, in order to successfully classify two very similar species in fine-grained, the most important thing is to find the discriminative part in the image that can distinguish the two species, and to be able to better represent the characteristics of these discriminative regional blocks.

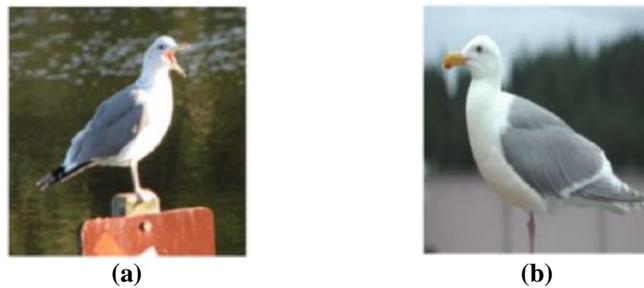

          (a)                                    (b)
**Fig.2** Example of fine-grained image classification. (a) California gull (b) Arctic gull.

Some existing data augmentation approaches, such as Hide and Seek [3], Random Erasing [1], GridMask [4], although they can achieve better occlusion in the face of coarse-grained images. However, they are likely to block important features in the face of fine-grained images.

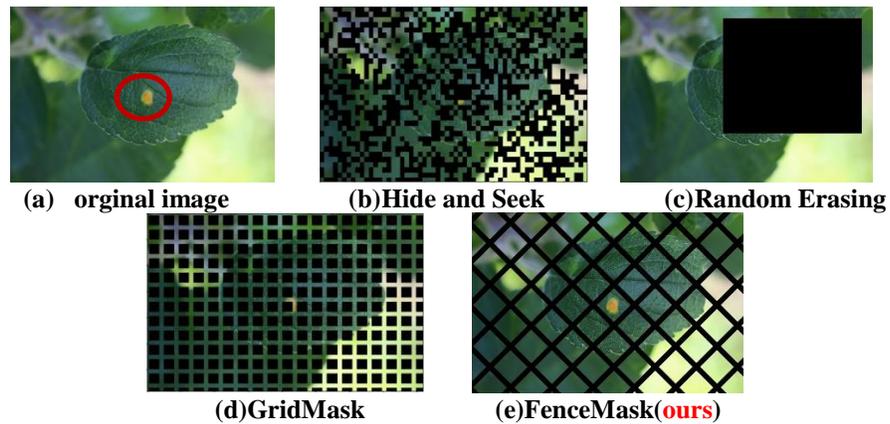

   **(a) orginal image**         **(b)Hide and Seek**         **(c)Random Erasing**

         **(d)GridMask**              **(e)FenceMask(ours)**

**Fig.3 Comparison of data augmentation approaches**

**Fig.3** shows the comparison results of the diseased leaves in Cornell leaves [1] under different data augmentation approaches. The yellow spots of the leaves are important features of the disease, while other data augmentation approaches completely erase the features, while our data augmentation method is easier to segment the features.

### 2.3 Small Object Detection

Object detection is aim to find the exact location of an object in a given image and mark out the category of the object. However, in the real scene, the size of the object varies greatly. Angle, posture and position of the object in the picture are different, they also have great influence on detection result. Besides, there may be overlap between the objects, which makes the object detection very difficult.

Meanwhile, small object detection is also a research issue in object detection. Most of the current object detection frameworks such as the classic one-stage approaches YOLO [13] and SSD [14], and two-stage approaches Faster-RCNN [15] were mainly designed for the general object dataset, the detection effect of them was not ideal for small object in the image.

For example, in the COCO dataset, objects with an area less than 32*32 will be considered as small objects. APs is used to count the mAP of these small objects. The results COCO dataset of current mainstream object detection algorithms is shown as **Table 1.** Whether two-stage method Faster R-CNN or one-stage method YOLO, SSD and anchor-free method CornetNet showed lower than average results in APs. It means that in Object detection task, small objects are often the bottleneck of model accuracy.

Using multi-scale feature map to predict different size objects is a common method to solve small object detection. In terms of feature integration, the previous research use skip connection or hyper-column to integrate high-resolution feature map to high-level semantic feature map, enhance its representational ability. Since multi-scale prediction approaches such as FPN [16] have become popular, many lightweight modules that integrate different feature pyramid have been proposed.

**Table 1.** Comparison of the accuracy of different object detectors on the MS COCO dataset.

| Method | AP | $AP_{50}$ | $AP_{75}$ | $AP_S$ |
|---|---|---|---|---|
| Faster R-CNN-FPN | 36.2% | 59.1% | 39.0% | 18.2% |
| YOLOv3-608 | 33.0% | 57.9% | 34.4% | 18.3% |
| SSD-VGG16-512 | 28.8% | 48.5% | 30.7% | 10.9% |
| RetinaNet-ResNet-50-500 | 32.5% | 50.9% | 34.8% | 13.9% |
| CornerNet-512 | 40.5% | 57.8% | 45.3% | 20.8% |

## 3 FenceMask

### 3.1 Method

FenceMask is an effective data augmentation method. Instead of randomly removing some pixels of image or generating rectangular block, it generates continuous blocking fences and fills in certain value. Unlike other method, the occlusion block of FenceMask has better sparsity and regularity, it helps to segment more features of object and reduce the total occlusion. The shape of our method looks like a fence, as shown in **Fig.3**. We use five parameters ($\boldsymbol{d_{min}}, \boldsymbol{d_{max}}, \boldsymbol{g_{min}}, \boldsymbol{g_{max}}$, F) to denote Fence.

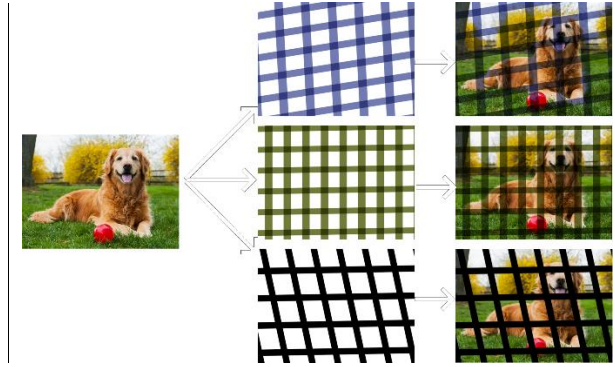

**Fig.3** This image shows how to achieve FenceMask by given five parameters($\boldsymbol{W_{min}}$, $\boldsymbol{W_{max}}, \boldsymbol{g_{min}}, \boldsymbol{g_{max}}$, F), multiply it with input image and eventually generate augmented image.

Assume that $W_x$ represents the width of horizontal fence, $W_y$ represents the width of vertical fence, $G_x$ represents the width of distance between adjacent fence in horizontal direction, and $G_y$ represents the width of distance between adjacent fence in vertical direction. A fence in both directions can be generated by ($W_x$, $W_y$) and ($G_x$, $G_y$) respectively and then multiply them to get full fence mask. In order to enrich the distribution of occlusion blocks and improve the randomness of it, we also introduce random rotation from 0 to 30 to the fence of each dimension. As for ($W_{min}$, $W_{max}$, $g_{min}$, $g_{max}$), we set a range within which these parameters will get the values at random, the max value of width is defined as $W_{max}$, the max value of width is defined as $W_{min}$, the max value of gap is defined as $G_{max}$, the min value of gap is defined as $G_{min}$.

$$G_x, G_y = random(G_{min}, G_{max})$$

$$W_x, W_y = random(W_{min}, W_{max})$$

In addition, F is a vector that represents filled value of fence in RGB channels, which is always set to zero. The pixel value distribution of augmented image s can be maintained by the value of F, especially when the occlusion block occupies a large proportion in the image, the information of the image can be protected by adjusting it. Like GridMask, our method can also set max probability of FenceMask, this probability linearly increases from 0 to max probability until the max epoch, and then keep is to end. To sum up, five parameters need to be set when using FenceMask, shown as follow.

$$(W_{min}, W_{max}, g_{min}, g_{max}, F)$$

Where $W_{min}$ and $W_{max}$ denote min value and max value of fence width, $G_{min}$ and $G_{max}$ denote min value and max value of gap between adjacent fence

## 4 Experiments

### 1 Dataset

We tested it on CIFAR10, CIFAR100 and ImageNet datasets for classification, COCO2017 and VisDrone dataset for detection, Oxford Flowers, Cornel Leaf and Stanford Dogs datasets for Fine-Grained Visual Categorization. Our method achieved significant performance improvement on Fine-Grained Visual Categorization task and VisDrone dataset.

### 2 Image Classification

The CIFAR10 dataset and CIFAR100 has 50,000 training images and 10,000 testing images. CIFAR10 has 10 classes, CIFAR10 has 100 classes, each has 5,000 training images and 1,000 testing images. We summarize the result on CIFAR10 in **Table 2.**

**Table 2.** Results of different data augmentation approaches on Classification.

| Model | Dataset | Baseline | GM | FM(ours) |
|---|---|---|---|---|
| Resnet18 | CIFAR10 | 95.28% | 95.92% | **96.12%** |
| Resnet18 | CIFAR100 | 76.68% | 76.78% | **77.10%** |

### 3 Fine-Grained Visual Categorization

It's a challenging task in fine-grained image recognition, and we're at Stanford Dogs [10], Oxford Flower 102 [9], Cornell Leaves [11], and compare our method with the current SOTA method.

The Stanford Dogs dataset has 120 categories and 20,580 images totally. Each class has approximately 150 images annotated with class labels and bounding boxes. The Oxford Flowers dataset has 2 version which has 17 categories and 102 categories separately. The 102 categories version is chosen as a widely used fine-grained image recognition benchmark. Each class consists of between 40 and 258 images. The Cornel Leaf dataset has 4 categories of apple orchard diseases leaves, each category consists of between 187 and 1399 images. This dataset is used in Plant Pathology Challenge for CVPR 2020-FGVC 7 workshop. Several challenges make this dataset more complex: imbalanced dataset of different disease categories, non-homogeneous background of images, images taken at different times of day, images from different physiological age of the plants, Multiple diseases in the same image, and different focus of image.

Using Resnet50 as our baseline model, we compare FenceMask with other SOTA data augmentation approaches, models are initialized by the same ImageNet pre-trained weights and then fine-tuned on the fine-grained dataset, the experimental results are shown in Table 4.

**Table 4.** Results of different data augmentation approaches on fine-grained datasets.

| Model | Dataset | Baseline | HaS | RE | GM | FM(ours) |
|---|---|---|---|---|---|---|
| Resnet50 | Stanford Dogs | 78.94% | 79.30% | 79.54% | 80.12% | **80.71%** |
| Resnet50 | Cornel Leaf | 96.54% | 97.24% | 96.87% | 97.54% | **98.04%** |
| Resnet50 | Oxford Flower | 92.46% | 92.61% | 92.91% | 93.12% | **93.31%** |

In the baseline experiment, we used the approaches of random left-right inversion, random lighting and random contrast adjustment. In the experiment, Adam optimizer was used, epoch was set to 50, and learning rate was 0.001. Compared with other models, our approaches have been significantly improved.

## 5  Discussion and Conclusion

We propose a novel data augmentation method named 'FenceMask' that exhibits outstanding performance in various computer vision tasks. It is based on the simulation of object occlusion strategy, which aim to achieve the balance between object occlusion and information retention of the input data. By enhancing the sparsity and regularity of the occlusion block, our augmentation method overcome the difficulty of small object augmentation and notably improve performance over baselines.

We tested its performance on a variety of computer vision tasks and demonstrated that our method has comparable performance on common datasets with current data enhancement methods, and particularly verified superior performance in fine-grained categorization and object detection datasets with many small objects.